\documentclass[runningheads]{llncs}
\usepackage{graphicx}
\usepackage{subcaption}
\usepackage{tikz}
\usepackage{comment}
\usepackage{amsmath,amssymb} % define this before the line numbering.
\usepackage{color}
\usepackage{algorithm}
\usepackage{algorithmic}
\usepackage{xspace}
\usepackage{multirow}
\usepackage{booktabs}

\usepackage[accsupp]{axessibility}  % Improves PDF readability for those with disabilities.
\begin{document}
% \renewcommand\thelinenumber{\color[rgb]{0.2,0.5,0.8}\normalfont\sffamily\scriptsize\arabic{linenumber}\color[rgb]{0,0,0}}
% \renewcommand\makeLineNumber {\hss\thelinenumber\ \hspace{6mm} \rlap{\hskip\textwidth\ \hspace{6.5mm}\thelinenumber}}
% \linenumbers
\pagestyle{headings}
\mainmatter
\def\ECCVSubNumber{6323}  % Insert your submission number here

\title{DistPro: Searching A Fast Knowledge Distillation Process via Meta Optimization} % Replace with your title

% INITIAL SUBMISSION 
%\begin{comment}
\titlerunning{ECCV-22 submission ID \ECCVSubNumber} 
\authorrunning{ECCV-22 submission ID \ECCVSubNumber} 
\author{Anonymous ECCV submission}
\institute{Paper ID \ECCVSubNumber}
%\end{comment}
%******************

% CAMERA READY SUBMISSION
% \begin{comment}
\titlerunning{Abbreviated paper title}
% If the paper title is too long for the running head, you can set
% an abbreviated paper title here
%
\author{Xueqing Deng\inst{1,3}\and
Dawei Sun\inst{2} \and
Shawn Newsam\inst{3} \and
Peng Wang \inst{1}}
\authorrunning{F. Author et al.}
% First names are abbreviated in the running head.
% If there are more than two authors, 'et al.' is used.
%
\institute{ByteDance Inc. \email{\{xueqingdeng,peng.wang\}@bytedance.com} 
\and ECE, UIUC, \email{\{daweis2\}@illinois.edu}
\and EECS, UC Merced, \email{\{snewsam\}@ucmerced.edu}}
% \end{comment}
%******************
\maketitle
% 1st describe issue 
% 2nd how we target the issue
% how are the results

\begin{abstract}
Recent Knowledge distillation (KD) studies show that different manually designed schemes impact the learned results significantly. Yet, in KD, automatically searching an optimal distillation scheme has not yet been well explored. In this paper, we propose \texttt{DistPro}, a novel framework which searches for an optimal KD process via differentiable meta-learning. Specifically, given a pair of student and teacher networks,  \texttt{DistPro} first sets up a rich set of KD 
connection from the transmitting layers of the teacher to the receiving layers of the student, and in the meanwhile, various 
transforms are also proposed for comparing feature maps along its pathway for the distillation. Then, each combination of a connection and a transform choice (pathway) is associated with a stochastic weighting process which indicates its importance at every step during the distillation. In the searching stage, the process can be effectively learned through our proposed bi-level meta-optimization strategy. In the distillation stage, \texttt{DistPro} adopts the learned processes for knowledge distillation, which significantly improves the student accuracy especially when faster training is required. Lastly, we find the learned processes can be generalized between similar tasks and networks. In our experiments, \texttt{DistPro} produces state-of-the-art (SoTA) accuracy under varying number of learning epochs on popular datasets, \textit{i.e.} CIFAR100 and ImageNet, which demonstrate the effectiveness of our framework.  
% Knowledge distillation (KD) is one of the core techniques to improve a neural network. 
% In this paper, we propose a novel meta-learning framework for finding an optimal knowledge distillation scheme given a pair of teacher and student neural networks. Here, we propose two key discoveries. 1) 
% During the distillation, both intermediate feature maps and outputs of the networks are adopted to distill the knowledge. 
% Specifically, we setup a rich set of pathways from transmitting layers of the teacher to receiving layers in the student, and different comparison strategies for these pathways are also proposed when computing loss between the corresponding feature maps. 
% Then, each pathway is associated with an importance factor, which can be effectively learnt through our differentiable meta-learning strategy. Finally, we adopt the learnt factors to do the knowledge distillation training. 
% In our case, a factor is not a fixed value after the learning, but defines a distillation process, which varies from begin to end of the distillation. 
% In our experiments, we show such a learnt process provides better distillation scheme than other state-of-the-art (SoTA) handcrafted alternatives in multiple vision tasks, including classification, segmentation and depth estimation, which demonstrates the effectiveness of our proposed framework.
\end{abstract}

\section{Introduction}
\label{sec:intro}
\vspace{-8pt}
\begin{figure}
    \centering
    \includegraphics[width=\textwidth]{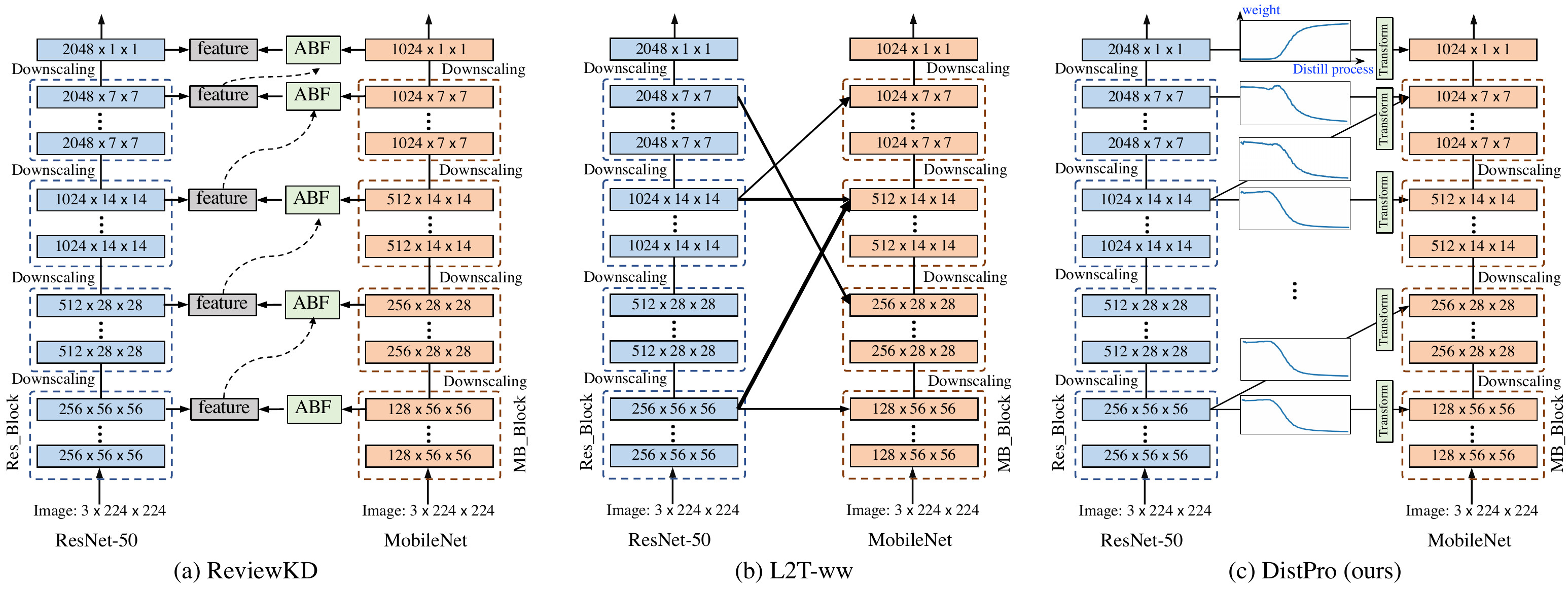}
    % \caption{Caption}
    \vspace{-18pt}
    \caption{Comparisons of distillation methods for learning process between teacher and student models. (a) Knowledge Review~\cite{chen2021distilling} proposes fixed sampled pathways by enumerating different configurations. (b) L2T-ww~\cite{jang2019learning}, adopts a meta-learning framework to learn a floating weight for each selected pathway. (c) Our framework learns distillation process for each pathway. }
    \vspace{-15pt}
    \label{fig:ovewview}
\end{figure}
% \begin{figure}
% \begin{subfigure}{0.32\textwidth}
%     \centering
%     \includegraphics[width=\textwidth]{figures/l2t-ww.pdf}
%     \caption{L2T-ww~\cite{jang2019learning}}
%     \label{fig:l2t}
%     \end{subfigure}
% \begin{subfigure}{0.32\textwidth}
% \centering
%     \includegraphics[width=\textwidth]{figures/reviewkd.pdf}
%     \caption{Knowledge Review~\cite{chen2021distilling}}
%     \label{fig:reviewkd}
% \end{subfigure}
% \begin{subfigure}{0.32\textwidth}
% \centering
%     \includegraphics[width=\textwidth]{figures/dispro.pdf}
%     \caption{DistPro (ours)}
%     \label{fig:distpro}
% \end{subfigure}

% \caption{Comparisons of distillation methods for learning process between teacher and student models. }
    
% \end{figure}

Knowledge distillation (KD) is proposed to effectively transfer knowledge from a well performing larger/teacher deep neural network (DNN) to a given smaller/ student network, where the learned student network often learns faster or performs better than that with a vanilla training strategy solely using ground truths. 

Since its first appearance in DNN learning~\cite{hinton2015distilling}, KD has achieved remarkable success in training efficient models for image classification \cite{zagoruyko2016paying}, image segmentation~\cite{liu2019structured}, object detection~\cite{chen2017learning}, \textit{etc}, contributing to its wide application in various model deployment over mobile-phones or other low-power computing devices~\cite{lyu2020differentially}.  Nowadays, KD has become a popular technique in industry to develop distilled DNNs to deal with billions of data per-day.

% large companies already deployed lots of distilled DNNs to deal with billions of data per-day.  % Therefore, small improvements over their accuracy can  bring strong benefits in many applications. 

% Most works are working on a single task, \eg, classification,  

To improve the distillation efficiency and accuracy, numerous handcrafted KD design schemes have been proposed, \textit{e.g.}, designing different distillation losses at outputs~\cite{liu2019structured,wang2020intra}, manually assigning intermediate features maps for additional KD guidance~\cite{zagoruyko2016paying,yim2017gift}. However, recent studies~\cite{gou2021knowledge,liu2020search,yao2021joint} indicate that the effectiveness of those proposed KD techniques is dependent on networks and tasks. Some recent works propose to search some configurations to conclude a better KD scheme. For instance, ReviewKD~\cite{chen2021distilling} (Fig.~\ref{fig:ovewview}(a)) proposes to evaluate a subset from a total of 16 pathways by enabling and disabling partial of them for image classification task. It comes to the conclusion that partial pathways are always redundant. However, our experimental results show that it does not hold to a semantic segmentation task, and after conducting a research of the pathways, we may obtain better results. 
L2T-ww~\cite{jang2019learning} (Fig.~\ref{fig:ovewview}(b)) takes a further step, it can not only set up multiple distillation pathways between feature maps, but also learn a floating weight for each pathway, which shows better performance than fixed weight.
%, instead it relies on the selected teacher and student networks. 
% Therefore, there remains a question how to choose proper intermediate guidance and what are their weight for distillation. 
This inspires us to explore deeper to find better KD schemes, and motivated by the learning rate scheduler, as illustrated in Fig.~\ref{fig:ovewview}(c), we brings in the concept of distillation process for each pathway $i$, \textit{i.e.} $\mathcal{A}^i = \{\alpha_t^i\}_{t=1}^{T}$,, where $T$ is the number of distillation steps. Thus, the importance of each pathway is dynamic and changes along the distillation procedure, which we find is beneficial in this paper.

However, searching a process is more difficult than finding a floating weight, which includes $T$ times more parameters. It is obviously non-practical to solve them via a brute force way. For example, randomly drawing a sample process, and performing a full training and validate of the network to evaluate its performance. 
Thanks to bi-level meta-learning~\cite{franceschi2018bilevel,liu2018darts}, we find our problems can be formulated and tackled in a similar manner for each proposed pathway during distillation. Such a framework not only skips the difficulty of random exploration through a valid meta gradient, but also naturally provides soft weighting that can be adopted to generate the process. Additionally, to effectively apply the framework and avoid  possible noisy gradients from the meta-training, we propose a proper normalization for each $\alpha_t = [\alpha^0_t, \cdot, \alpha^N_t]$, where $N$ is number of pathways. We call our framework \texttt{DistPro} (Distillation Process). % and in \figref{fig:ovewview}, we compare \texttt{DistPro} with other similar KD schemes with intermediate pathways to distinguish our contribution in this paper. 
In our experiments, we show that \texttt{DistPro} produces better results on various tasks, such as classification with CIFAR100~\cite{krizhevsky2009cifar} and ImageNet1K~\cite{imagenet_cvpr09}, segmentation with CityScapes~\cite{Cordts2016Cityscapes} and depth esimation with NYUv2~\cite{Silberman:ECCV12}. 

Finally, % we found the weights of pathways from low-level feature maps of the teacher networks are relatively large at the beginning while turns to be small at the end. While the behavior of the weights from high-level features is reversed. This indicates that an optimal routine for distillation could be that we should let the student network learn simple knowledge at early stage, and difficult knowledge at later stage. 
we find our learned process remains similar with minor variation across different network architectures and tasks as long as it uses the same proposed pathways and transforms, which indicates that the process can be generalized to new tasks. In practice, we transfer the process learned by CIFAR100 to ImageNet1K, and show that it improves over the baselines and accelerates the distillation (2x faster than ReviewKD~\cite{chen2021distilling} as shown in Tab.~\ref{tab:transfer}).

% In addition, for a given pathway of distillation, our meta-learning framework is able to generates a weighting process, which actually contains more information beyond the final learnt value. 
% First, during the training, the importance weights to learn are located at loss terms rather than in the architecture, therefore, 
% Second, the weights are not a fixed value 

% In summary, as illustrated in \figref{fig:ovewview} and \figref{fig:algo}, DistPro works as follows. For a given pair of teacher and student networks, we first sample and define a set of transmitting feature maps from the teacher and receiving feature maps from the student. Meanwhile, a set of transforms are also proposed, which converts a transmitting feature map to match with a receiving feature map for loss computation, yielding a distillation pathway. For each pathway, an importance factor $\alpha$ is assigned. Then, after preforming our meta-learning step, a learnt process for $\alpha$ is stored. Finally, we adopt the process to reweighting each pathway for our KD training, generating a distilled student model for deployment. We validate our approach based on various vision tasks including image classification, image segmentation and depth estimation. In all experiments, our strategy performs better than other SoTA KD methods specifically designed for each task~\cite{chen2021distilling, wang2020intra}, which demonstrates its effectiveness.

In summary, our contributions are three-folds 1) We propose a meta-learning framework for KD, \textit{i.e.} DistPro, to efficiently learn an optimal process to perform KD.  2) We verify \texttt{DistPro} over various configurations,  architecture and task settings, yielding significant improvement over other SoTA methods. 3) Through the experiments, we find processes that can generalize across tasks and networks, can potentially benefit KD in new tasks without additional searching. Our code and implementation will be released upon the publication of this paper. %Rather than adopting a fixed distillation weights to balance different losses, we found using a weighting process learnt by LATTE produces better results in our experiments, which potentially could be generalized to other learning scenarios. 

\section{Related Works}
\vspace{-10pt}
% in the domain of KD, how we are located and different from previous approaches
% in the other related works, meta-learning for few-shorts, Darts, hyper-parameter tunning, how we are related.
% {\def\ie{\latinabbrev{i.e }}}
\textbf{Knowledge distillation.} 
Starting under the name knowledge transfer~\cite{bucilua2006model,urner2011access}, knowledge distillation (KD) is later popularized owing to Hinton et.al~\cite{hinton2015distilling} for training efficient neural networks. Thereafter, it has been a popular field in the past few years, in terms of designing KD losses~\cite{wang2018kdgan,wang2020intra}, combination with multiple tasks~\cite{oord2018parallel,devlin2018bert} or dealing with specific issues, eg few-shot learning~\cite{kimura2018few}, long-tail recognition~\cite{xiang2020learning}. Here, we majorly highlight the works that are closely related to ours, in order to locate our contributions.
% In KD, a student model is trained using the supervision of a teacher model. In other words, knowledge in a teacher model is distilled into a student model.

According to a recent survey~\cite{gou2021knowledge}, current KD literature include multiple knowledge types, eg response-based~\cite{muller2019does}, feature-based~\cite{Romero2015FitNetsHF,heo2019comprehensive} and relation-based~\cite{tung2019similarity} knowledge. In addition, in terms of distillation method, we have offline~\cite{hinton2015distilling}, online~\cite{chen2020online} and self-distillation~\cite{zhang2019your}. For distillation algorithms, various distillation criteria are proposed such as adversarial-based~\cite{wang2018kdgan}, attention-based~\cite{huang2017like}, graph-based~\cite{lee2019graph} and lifelong distillation~\cite{chen2018lifelong} etc.  Finally, based on a certain task, KD can also extend with different task-aware metrics, eg speech~\cite{oord2018parallel}, NLP~\cite{devlin2018bert} etc. 
In our principle, we hope all the surveyed KD schemes, ie knowledge types, methods under certain task settings, can be pooled with a universal way to a search space, in order to find the best distillation.
While in this paper, we take the first step towards this goal by exploring a sub-field in this whole space, which is already a challenging problem to solve. 
Specifically, we adopt the setting of offline distillation with feature-based and response-based knowledge, where both network responses and intermediate feature maps are adopted for KD. 
For KD method, we use attention-based methods to compare feature responses, and apply the KD model to vision tasks including classification, segmentation and depth estimation

Inside this field, knowledge review~\cite{chen2021distilling} and L2T-ww~\cite{jang2019learning} are the most related to our work. The former investigates the importance of a few pathways and propose a knowledge review mechanism with a novel connection pattern, ie, residual learning framework. It provides SoTA results in several commonly comparison benchmarks. The latter learns a fixed weights for a few intermediate pathways for few-shot knowledge transfer. As shown in Fig.~\ref{fig:ovewview}, \texttt{DistPro} finds a distillation process. Therefore, we extend the search space. In addition, for dense prediction tasks, one related work is IFVD~\cite{wang2020intra}, which proposes an intra-class feature variation comparison (IFVD).  \texttt{DistPro} is free to extend to dense prediction tasks,  and it also obtain extra benefits after combined with IFVD.
% Our work extends \cite{chen2021distilling} by searching for optimal pathways and feature transform architectures. 

% 1) learning hyperparameters, 2) architectures, 3) tasks 
% talked about other, KD + Meta-learning, for few-shot, importance weight etc. 
\noindent\textbf{Meta-learning for KD/hyperparameters.}  % here explain why we use gradient based meta-learning scheme
To automate the learning of a KD scheme, we investigated a wide range of efficient meta-learning methods in other fields that we might adopt. 
For examples, L2L~\cite{andrychowicz2016learning} proposes to learn a hyper-parameter scheduling scheme through a RNN-based meta-network.  Franceschi et.al~\cite{franceschi2018bilevel} propose an gradient-based approach without external meta-networks. Later, these meta-learning ideas have also been utilized in tasks of few-shot learning (eg learning to reweight~\cite{ren2018learning}),  learning cross-task knowledge transfer(eg learning to transfer~\cite{jang2019learning}), and neural architecture search (NAS)(eg DARTS~\cite{liu2018darts}).  Through these methods share similar framework, while it is critical to have essential embedded domain knowledge and task-aware adjustment to make it work. In our case, inspired by these methods, we majorly utilize the gradient-based strategy due to its efficiency for KD scheme learning, and also first to propose using the learnt process additional to the learnt importance factor. 

Finally, knowledge distillation for NAS has also drawn significant attention recently. 
%The motivation of AKD is that the teacher’s architecture has to match with the student to facilitate knowledge transfer. 
For example, Liu et.al~\cite{liu2020search} try to find student models that are best for distilling a given teacher, while Yao et.al~\cite{yao2021joint} propose to search architectures of both the student and the teacher models based on a certain distillation loss. Though different from our scenario, \textit{i.e.} fixed student-teacher architectures, it raises another important question of how to find the Pareto optimal inside the union space of architectures and KD schemes under certain resource constraint, which we hope could inspire future researches.

\begin{figure}[t]
    \centering
    \includegraphics[width=\textwidth]{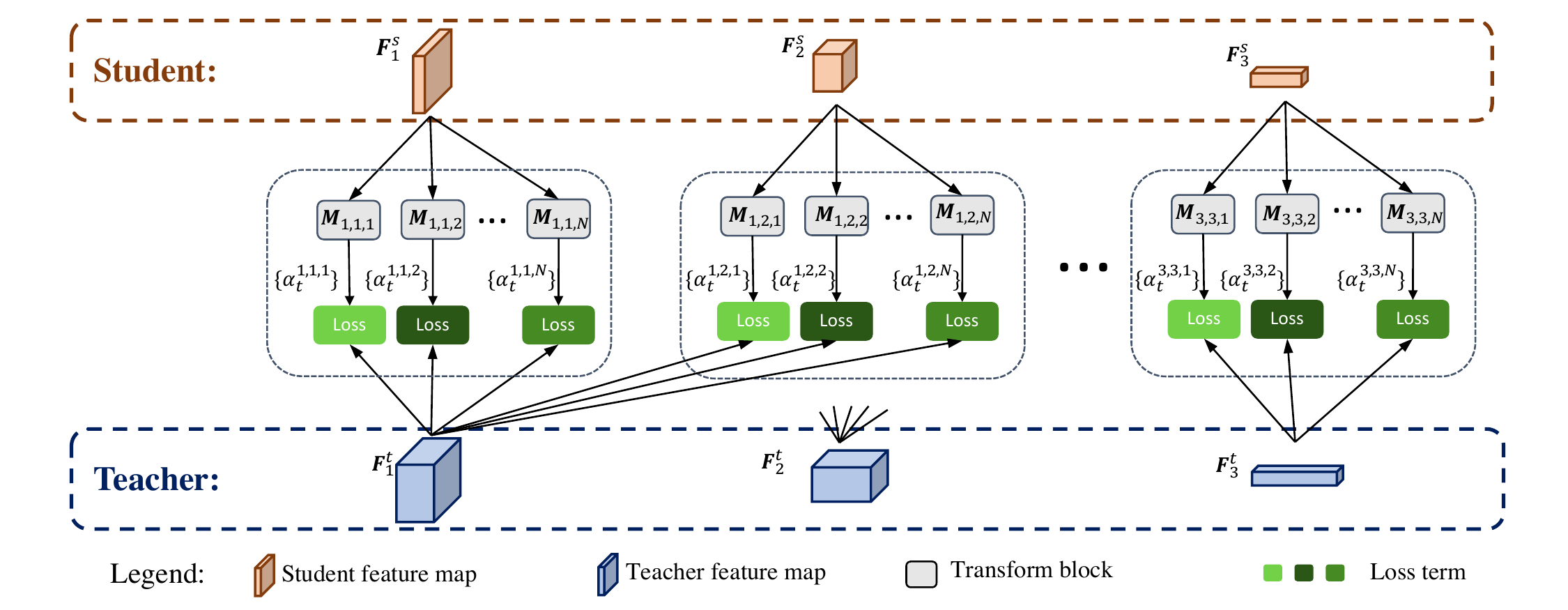}
    \vspace{-15pt}
    \caption{Illustration of the search space. Two groups of feature maps are selected from the student and the teacher. For each pair of feature maps $\mathcal{F}_j^s$ and $\mathcal{F}_i^t$, we insert $N$ candidate transform blocks $M_{i,j,1}, M_{i,j,2}, \cdots, M_{i,j,N}$ and get $N$ loss terms. For each loss term, a distill process $\mathcal{A}^{i,j,k} = \{\alpha^{i,j,k}_t\}_{t=1}^{T}$ is assigned.}
    \vspace{-10pt}
    \label{fig:arch}
\end{figure}

\section{Approach}
\label{sec:algo}

In this section, we elaborate \texttt{DistPro} by first setting up the KD pathways with intermediate features which constructs our search space, establishing the notations and definition of our KD scheme. Then, we derive the gradient, which is used to generate our process for the KD scheme. At last, the overall algorithm is presented.

\vspace{-5pt}
\subsection{KD with intermediate features}\label{sec:prelim}
\vspace{-5pt}
Numerous prior works~\cite{ji2021feature,chen2021distilling} have demonstrated that intermediate feature maps from neural network can benefit distilling to  student neural network. Motivated by this, we design our approach using feature maps.
Let the student neural network be $\mathcal{S}$ and the teacher neural network be $\mathcal{T}$. Given an input example $\mathbf{X}$, the output of the student network is as follows

\begin{equation}
\mathcal{S}(\mathbf{X}) := \mathcal{S}_{L_s} \circ \cdots \circ \mathcal{S}_{2} \circ \mathcal{S}_{1}(\mathbf{X}),
\end{equation}

where $\mathcal{S}_i$ is the $i$-th layer of the neural network and $L_s$ is the number of layers. The $k$-th intermediate feature map $\mathcal{F}^s_k$ of the student is defined as follows,

\[
\mathcal{F}^s_k(\mathbf{X}) := \mathcal{S}_{k} \circ \cdots \circ \mathcal{S}_{2} \circ \mathcal{S}_{1}(\mathbf{X}),~1 \leq k \leq L_s.
\]

% Please also note that we might drop some pathways of $\mathbf{X}$ when they are clear from context. 
Similarly, the feature map of the teacher is denoted by $\mathcal{F}^t_k$, $1 \leq k \leq L_{te}$.

Knowledge can be distilled from a pathway between $i$-th feature map of the teacher and $j$-th feature map of the student by penalizing the difference between these two feature maps, ie, $\mathcal{F}^t_i$ and $\mathcal{F}^s_j$. Since the feature maps may come from any stage of the network, they may not be in the same shape and thus not directly comparable. Therefore, additional computation are required to align these two feature maps to the same shape. To this end, a transform block is added after $\mathcal{F}^s_j$, which could be in any forms of differentiable computation. In our experiments, the transform block consists of multiple convolution layers and an interpolation layer to align the spatial and channel resolution of the feature map. Denoting the transform block by $\mathcal{M}$, then the loss term measuring the difference between these two feature maps is as follows,
\[
\ell(\mathcal{F}^s_j, \mathcal{F}^t_i) := \delta(\mathcal{M}(\mathcal{F}^s_j), \mathcal{F}^t_i),
\]
where $\delta$ is the optional distance function, which could be L1 distance or L2 distance as used in~\cite{ji2021feature}, etc.

\subsection{The Distillation Process}
\vspace{-5pt}
Now we will be able to build pathways to connect any feature layer of the teacher to any layer of the student with appropriate transforms. However, as discussed in Sec.~\ref{sec:intro}, not all of these pathways are beneficial. 
% For instance, only the transfers from lower levels of teacher to deeper levels of students are useful for image classification on CIFAR100~\cite{chen2021distilling}.  
This motivates us to design an approach to find out the importance of each possible pathway by assigning an importance factor for it. Different from the existing work~\cite{jang2019learning,chen2021distilling}, the importance factor here is a process. 
Formally, a stochastic weight process $\mathcal{A}^i = \{\alpha^i_t\}_{t=1}^{T}$ is associated with the pathway $i$, where $T$ is the total learning steps. Here, $\alpha^i_t$ describes the importance factor at different learning step $t$. % Such learned distillation process have multiple advantages, in particular it can benefit faster distillation shown in our experimental results (~\secref{sec:exp_imagenet}).
% H  an interesting question to ask is how to find the optimal KD scheme. Next, we formulate this as an optimization problem.

\begin{figure}[!t]
    \centering
    \includegraphics[width=\linewidth]{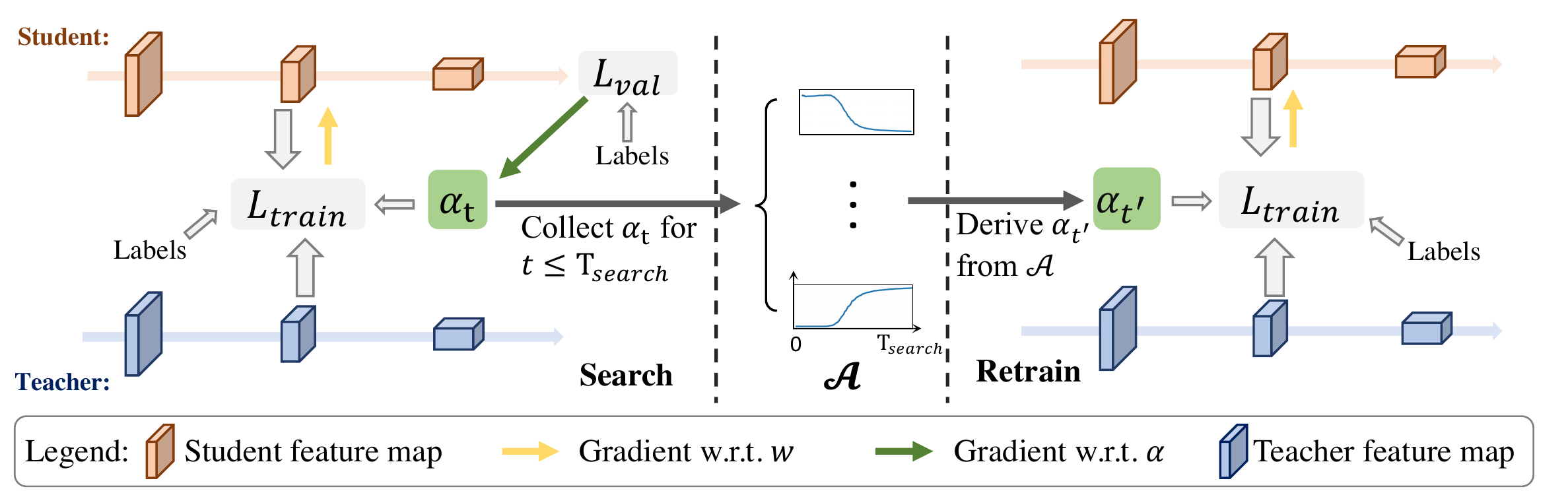}
    \vspace{-15pt}
    \caption{Two phases of the proposed algorithm. In the search phase, student $\theta$ and $\alpha_t$ are computed. 
    In the retrain phase, the learned process of $\mathcal{A}=\{\alpha_t,  0\le t \le T_{search}\}$ is interpolated to be used for KD, and only $\theta$ is updated.}
    \vspace{-10pt}
    \label{fig:algo}
\end{figure}

Let $D_{train} := \{(\mathbf{X}_i, y_i)\}_{i=1}^{|D_{train}|}$ and $D_{val} := \{(\mathbf{X}_i, y_i)\}_{i=1}^{|D_{val}|}$ be the training set and the validation set respectively, where $y_i$ is the label of sample $\mathbf{X}_i$. We assume that for each pair of feature maps, $\mathcal{F}^t_i$ and $\mathcal{F}^s_j$, we have $N$ candidate pre-defined transforms, $\mathbf{M}_{i,j,1}, \mathbf{M}_{i,j,2}, \cdots, \mathbf{M}_{i,j,N}$. The connections together with the transforms construct our search space as shown in Fig.~\ref{fig:arch}.

We now define the search objective which consists of the optimizations of the student network and $\mathcal{A}$. The student network is trained on the training set with a loss encoding the supervision from both the ground truth label and the neural network. Specifically, denoting the parameters of the student and the transforms by $\theta$, the loss on the training set at learning step $t$ is defined as follows,

\begin{equation}
\begin{aligned}
L_{train}(\theta_t, \alpha_t) &= \frac{1}{|D_{train}|} \sum_{(\mathbf{X}, y) \in D_{train}} \Big(\delta_{label}(\mathcal{S}(\mathbf{X}), y)& \\
&~~~~~~~~~~~~~~~~+ \sum_{i=1}^{L_{te}} \sum_{j=1}^{L_s} \sum_{k=1}^{N} \alpha_t^{i,j,k} \delta\left(\mathbf{M}_{i, j, k}\left(\mathcal{F}^s_j(\mathbf{X})\right), \mathcal{F}^t_i(\mathbf{X})\right)\Big),\\
\alpha_t^{i,j,k} &= \frac{\exp(\tilde{\alpha}_t^{i,j,k})}{\sum_{i, j, k}\exp(\tilde{\alpha}_t^{i,j,k}) + \exp(g)}
\label{eq:alphatrain}
\end{aligned}
\end{equation}

where $\alpha_t^{i,j,k} \in \mathbb{R}_{\geq 0}^{L_{te} \times L_s \times N}$ are the importance factors at training step t and $\delta_{label}$ is a distance function measuring difference between predictions and labels. Here, $\alpha_t = [\alpha_t^{0, 0, 0}, \cdots]$ controls the importance of all knowledge distill pathway at training step $t$. For numerical stability and avoiding noisy gradients, we apply a biased softmax normalization with parameter $g=1$ to compute $\alpha_t$ which is validated from various normalization strategies in our experiments. This is a commonly adopted trick in meta-learning for various tasks such as NAS~\cite{xie2018snas,liu2018darts} or few-shot learning~\cite{ren2018learning,jang2019learning}. 
%i.e., the decision variable of the optimization is actually $\tilde{\alpha}_t$, and $\alpha_t$ is obtained by normalizing $\tilde{\alpha}_t$. 
More details about the discussion of normalization methods can be found in the experimental section (Sec.~\ref{sec:ablation}).

Next, our goal is to find an optimal sampled process $\mathcal{A}^*$ yielding best KD results, where the validation set is often used to evaluate the performance of the student trained on unseen inputs. To this end, following~\cite{liu2018darts}, the validation loss is adopted to evaluate the quality of $\mathcal{A}$, which is defined as follows,
\[
\setlength\abovedisplayskip{-1pt}
L_{val}(\theta) := \frac{1}{|D_{val}|} \sum_{(\mathbf{X}, y) \in D_{val}} \delta_{label}(\mathcal{S}(\mathbf{X}), y).
\vspace{-5pt}
\]

Finally, we formulate the bi-level optimization problem over $\mathcal{A}$ and network parameter $\theta$ as:
\begin{equation}
\setlength\abovedisplayskip{-15pt}
\begin{aligned}
\min_{\mathcal{A}} \quad & L_{val}\left(\theta^*(\mathcal{A})\right) \\ 
\textrm{s.t.} \quad & \theta^*(\mathcal{A}) = \arg\text{f}_{\theta} \quad L_{train}(\Theta, \mathcal{A}).
\label{eq:bilevel}
\end{aligned}
\vspace{-5pt}
\end{equation}
where $\theta^*(\mathcal{A})$ is the parameters of the student neural network trained with the loss process defined with $\mathcal{A}$, \textit{i.e.} $L_{train}(\Theta, \mathcal{A}) = \{L_{train}(\theta_t, \alpha_t)\}_{t=1}^{T}$. The ultimate goal of the optimization is to find an $\mathcal{A}$ such that the loss on the validation set is minimized.% Here $\argf$ is a function to extract the parameter  % different from nas
Note here, similar problem has been proposed in NAS~\cite{xie2018snas,liu2018darts} asking for a fixed architecture, but our problem is different and harder, and can be a reduced to theirs if we enforce all the value in $\mathcal{A}$ to be the same. However, similarly, we are able to apply gradient-based method following the chain-rule to solve this problem. %In the next section, we derive an approximation of the gradient.

\begin{algorithm}[tb]
\caption{DistPro}
\label{alg:algorithm}
\textbf{Input}: Full train data set $D$; Pre-trained teacher; Initialization of $\alpha_0$, $\theta_0$; Number of iterations $T_{search}$ and $T$.\\
\textbf{Output}: Trained student.
\begin{algorithmic}[1] %[1] enables line numbers
\STATE Split $D$ into $D_{train}$ and $D_{val}$.
\STATE Let $t=0$.
\WHILE{$t < T_{search}$}
\STATE Compute $\alpha_t$ with descending the gradient approximation in Eq.~\ref{eq:final} and do normalization.
\STATE Update $\theta_t$ by descending $\nabla_\theta L_{train}(\theta_t; \alpha_t)$ in Eq.~\ref{eq:alphatrain}.
\STATE Push the current $\alpha_t$ to $\mathcal{A}$.
\ENDWHILE
\STATE Interpolate $\mathcal{A}$ with length of $T$.
\STATE Set $D_{train} = D$ or use a new $D_{train}$ in another task.
\STATE Reset $t = 0$.
\WHILE{$t < T$}
\STATE Load the $\alpha_t$ corresponding to the current $t$. \label{algo:line:load}
\STATE Update $\theta$ by descending $\nabla_\theta L_{train}(\theta, \alpha_t)$.
\ENDWHILE
\end{algorithmic}
\end{algorithm}

\subsection{Learning the Process $\mathcal{A}$}
\label{sec:meta-learning}
% smooth assumption, and theta dependency
\vspace{-5pt}
From the formulation in Eq.~\ref{eq:bilevel}, directly solving the issue is intractable. Therefore, we propose two assumptions to simplify the problem. First, smooth assumption which means the next step of the process should be closed to previous one. This is commonly adopted in DNN training with stochastic gradient decent (SGD) with a learning rate scheduler~\cite{kingma2014adam}. Second, similar learning procedure assumption, which means among different distillation training given a teacher/student pair, at same step $t$, the student will have similar parameter $\theta_t$. This assumption allows us to search for $\mathcal{A}$ with a single training procedure, which also holds from our experiments when batch size is relatively large (\textit{e.g.} 512 for ImageNet).

Based on these assumptions, we let $\alpha_{t+1} = \alpha_{t} + \gamma\Delta(\alpha_t | \theta_t)$, where $\gamma$ controls its changing ratio to be small.  Then, we are able to adopt a greedy strategy to break the original problem of Eq.~\ref{eq:bilevel} down to a sequence of single steps of optimization, which can be defined as, 
\begin{equation}
\setlength\abovedisplayskip{-2pt}
\begin{aligned}
    \alpha_{t+1} = \alpha_t - \gamma\nabla_{\alpha} L_{val}\left(\theta_{t+1}(\alpha_t)\right)\\  \nonumber
    \textrm{s.t} ~~ \theta_{t+1}(\alpha_t) = \theta_t - \xi \nabla_\theta L_{train}(\theta_t, \alpha_t)
\label{eq:singlestepapprox}
\end{aligned}
\end{equation}
where $\xi$ is the learning rate of inner optimization for training student network. In practice, the inner optimization can be solved with more sophisticated gradient-based method, eg, gradient descent with momentum. In those cases, Eq.~\ref{eq:singlestepapprox} has to be modified accordingly, but all the following analysis still applies.

% As mentioned earlier, computing the gradient with respect to $\alpha$, i.e., $\nabla_{\alpha} L_{val}\left(w^*(\alpha)\right)$, is intractable due to the nested structure of the optimization problem. Inspired by previous work including Differentiable Architecture Search~\cite{liu2018darts} and meta-learning~\cite{finn2017model}, we approximate the gradient as follows.

% At this stage, instead of computing gradient at the exact optimum $w^*(\alpha)$ of the inner optimization, we compute it at the result of the single-step gradient descent,
% \begin{equation}
% \begin{aligned}
% & \nabla_{\alpha} L_{val}\left(w^*(\alpha)\right)\\
% \approx & \nabla_{\alpha} L_{val}\left(w - \xi \nabla_w L_{train}(w, \alpha)\right).
% \end{aligned}
% \label{eq:singlestepapprox}
% \end{equation}
% Here, $w$ is the current parameters of the neural network and $\xi$ is the learning rate of the inner optimization. We approximate $w^*(\alpha)$ with a single step of gradient descent from the current parameter $w$. Please note that in practice, the inner optimization might be solved with more sophisticated gradient-based method, \eg, gradient descent with momentum. In those cases, Eq.~\equref{eq:singlestepapprox} has to be modified accordingly, but all the following analysis still applies.

Next, we apply the chain rule to Eq.~\ref{eq:singlestepapprox} and get
\begin{equation}
\begin{aligned}
\nabla_{\alpha} L_{val}\left(\theta_t - \xi \nabla_\theta L_{train}(\theta_t, \alpha_t)\right)
= - \xi \nabla_{\alpha, \theta }^2 L_{train}(\theta_t, \alpha_t) \nabla_\theta L_{val}(\theta_{t+1}),
\end{aligned}
\end{equation}
However, the above expression contains second-order derivatives, which is still computational expensive. Next, we approximate this second-order derivative with finite difference as introduced in~\cite{liu2018darts}. Let $\epsilon$ be a small positive scalar and define the notation $\theta^\pm = \theta \pm \epsilon \nabla_\theta L_{val}(\theta_{t+1})$. Then, we have
\begin{equation}
\begin{aligned}
\nabla_{\alpha, \theta}^2 L_{train}(\theta_t, \alpha_t) \nabla_\theta L_{val}(\theta_{t+1})
=  \frac{\nabla_{\alpha}L_{train}(\theta^+, \alpha_t) - \nabla_{\alpha}L_{train}(\theta^-, \alpha_t)}{2\epsilon}.
\label{eq:final}
\end{aligned}
\end{equation}
Finally, we set an initial value $\alpha_0 = 1$ and launch the greedy learning procedure. Once learned, we push all computed $\alpha_t$ with $T$ steps to a sequence, which is served as a sample of learned stochastic distillation process $\mathcal{A}$, and it can be used for retraining the student network with KD in the same or similar tasks. 
% To conclude, we get the following approximation
% \begin{equation}
% \begin{aligned}
% & \nabla_{\alpha} L_{val}\left(w^*(\alpha)\right)\\
% \approx & - \xi \frac{\nabla_{\alpha}L_{train}(w^+, \alpha) - \nabla_{\alpha}L_{train}(w^-, \alpha)}{2\epsilon}.
% \end{aligned}
% \label{eq:final}
% \end{equation}
\vspace{-5pt}
\subsection{Acceleration and adopting $\mathcal{A}$ for KD.} 
\vspace{-5pt}
Now let us take a closer look at the approximation. To evaluate the expression in Eq.~\ref{eq:final}, the following items have to be computed. First, computing $\theta_{t+1}$ requires a forward and backward pass of the student, and a forward pass of the teacher. Then, computing $\theta^\pm$ requires another forward and backward pass of the student. Finally, computing $\nabla_{\alpha}L_{train}(\theta^{\pm}, \alpha_t)$ requires two forward passes of the student. % Please note that the gradient of $L_{train}$ with respect to an element of $\alpha$ is just the feature map loss corresponding to this element, so no further backward pass of the student is needed. 
In conclusion, evaluating the approximated gradient in Eq.~\ref{eq:final} entails one forward pass of the teacher, and four forward passes and two backward passes of the student in total. This is a time consuming process, especially when the required KD learning epoch is large, \textit{e.g.} $\geq 100$. 

In meta-learning NAS literature, to avoid $2_{nd}$ order approximation, researchers commonly adopt $1_{st}$ order training~\cite{xie2018snas,liu2018darts} solely based on training loss. However, this is not practical in our case since the hyperparameters are defined over training loss.  $1_{st}$ order gradient over $\alpha_t$ will simply drive all value to $0$. Therefore, in our case, we choose to reduce the learning epochs of $\mathcal{A}$ to $T_{search}$, which is much smaller than $T$, and then expand it to a sequence with length $T$ using linear interpolation. The choices of $T_{search}$ can be dynamically adjusted based on the dataset size, which will be elaborated in Sec.~\ref{sec:exp}.  % In our experiment, $T_{search}$ is reduced to $T/4$, and we find it impacts little for the performance.  

\begin{figure}[t]
    \centering
    \begin{subfigure}{0.49\textwidth}
     \includegraphics[width=\textwidth]{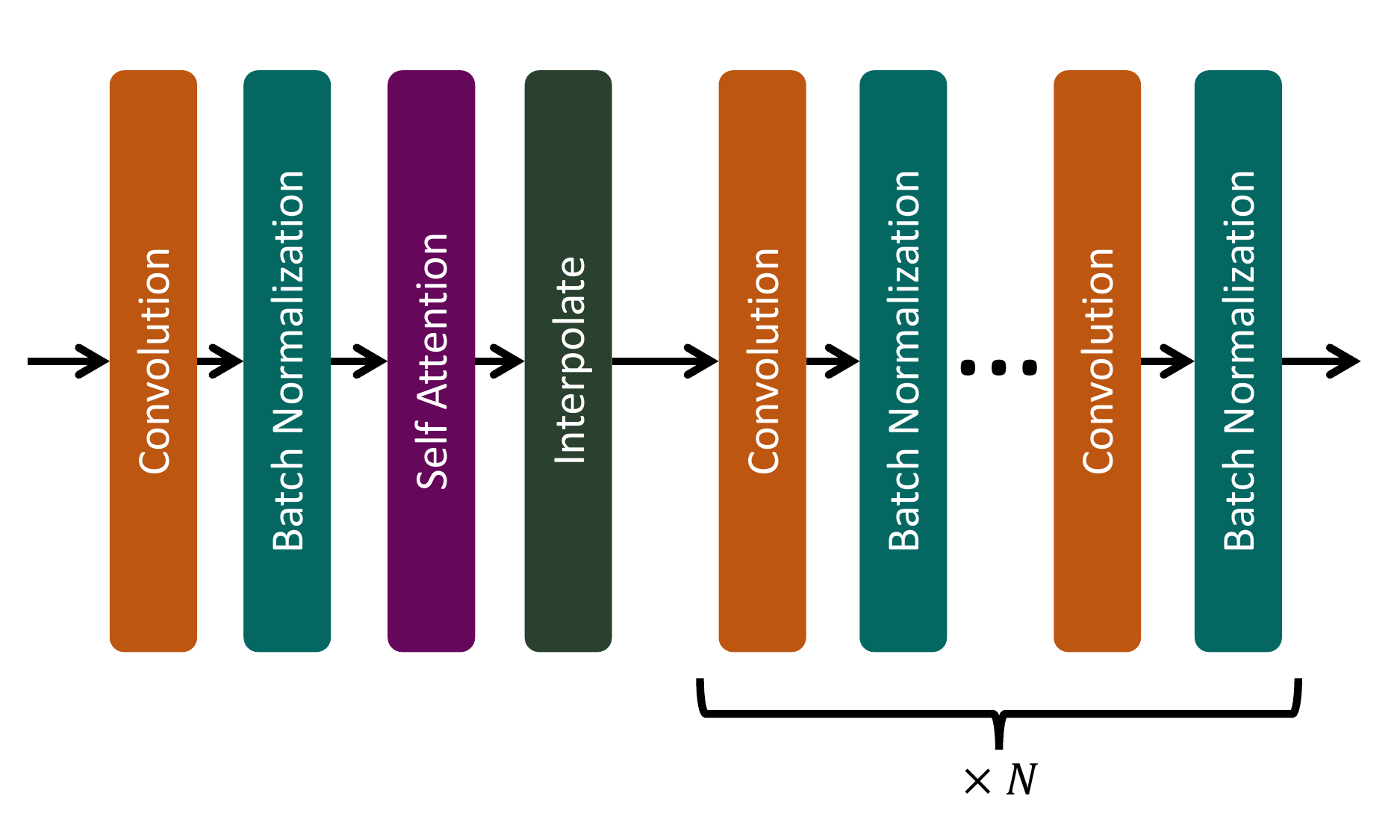}
    \caption{The architecture of the transform block. The self-attention block is illustrated in Fig.~\ref{fig:selfatt}.}
    \label{fig:trans}
    \end{subfigure}
    \begin{subfigure}{0.49\textwidth}
      \includegraphics[width=\textwidth]{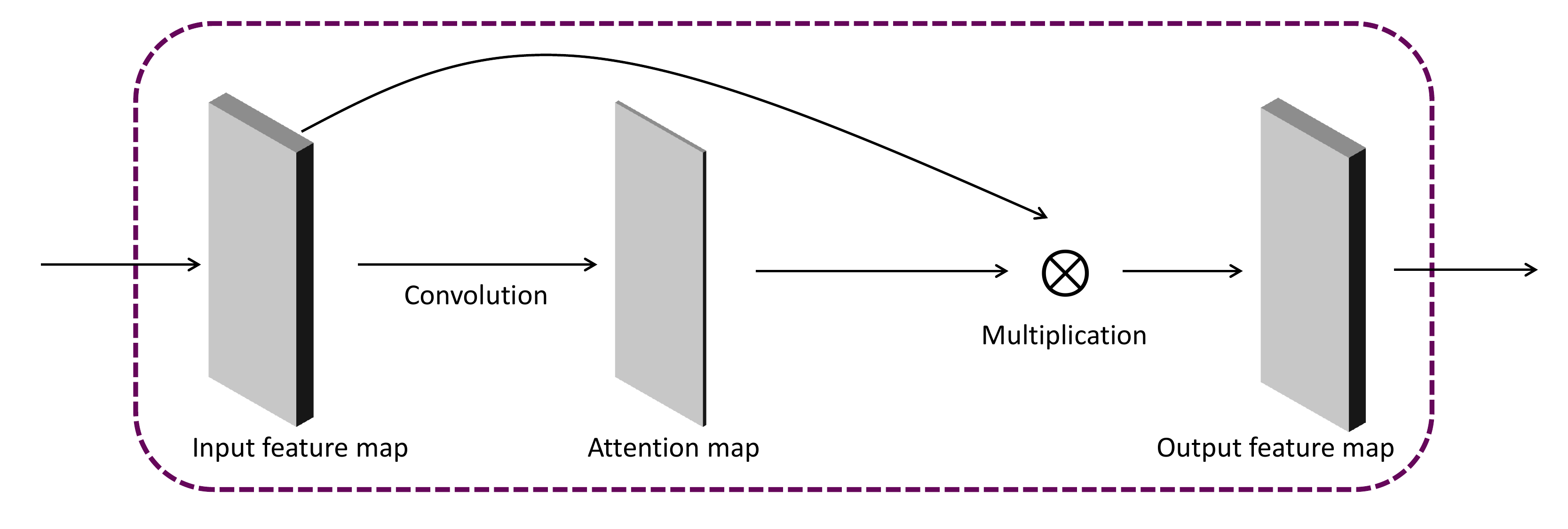}
    \caption{The architecture of the self-attention block. A convolution layer is applied to the input feature map to generate a $1$-channel attention map. Then the input feature map is multiplied with the attention map.}
    \label{fig:selfatt}
    \end{subfigure}
    \vspace{-15pt}
   \caption{Transform blocks}
   \vspace{-10pt}
\end{figure}

Additionally, since we have more pathways than other KD strategy~\cite{chen2021distilling}, therefore, the loss computation cost can not be ignored. To reduce the cost, at step $t$, we use a clip function to all $\tilde{\alpha}_t$ in Eq.~\ref{eq:alphatrain} with a threshold $\tau = 0.5$, and drop corresponding computation when $\tilde{\alpha}_t \leq \tau$. This can save us $60\%$ of loss computational cost in average, resulting in comparable KD time with our baselines.

In summary, the overall \texttt{DistPro} process is presented in Alg.~\ref{alg:algorithm}, where two phases of the algorithm are explained in order.
The first phase of the algorithm is for searching the scheme $\mathcal{A}$. To this end, $\alpha_t$ and $\theta_t$ are computed alternately. The $\alpha_t$ obtained in each step is stored for future usage.
The Second phase of the algorithm is for retraining the neural network with all the available training data and the searched $\mathcal{A}$. %There are two options for how to use the searched $\alpha$. % One can use just the $\alpha$ obtained at the last iteration in the searching phase for each iteration of the retraining phase. However, as will be shown in the experimental section, the evolution of $\alpha$ in the searching phase encodes much richer information. 
% One could make use of this information by using different $\alpha$ at each iteration of the retraining phase. To this end, we load a new value from the stored $\alpha$'s in Line~\ref{algo:line:load}. Since $N_{search} \neq N_{retrain}$, we use linear interpolation to compute the $\alpha$ for each iteration in the retraining phase. Specifically, the $\alpha$ for iteration $t$ is $i$-the stored $\alpha$, where $i = \lfloor\frac{t N_{search}}{N_{retrain}}\rfloor$.

% + clipping for saving computation, and a clipping function with threshold $\tau=0.01$ 

% normalization Done
% table 1 Done
% alphas Done
% figure 1 caption feature map Done
% illustration for algo Done

\vspace{-3pt}
\section{Experiments}
\label{sec:exp}

\vspace{-8pt}
% \begin{figure}[tbhp]
%     \centering
%     \includegraphics[width=\linewidth]{figures/selfatt.pdf}
%     \caption{The architecture of the self-attention block. A convolution layer is applied to the input feature map to generate a $1$-channel attention map. Then the input feature map is multiplied with the attention map.}
%     \label{fig:selfatt}
% \end{figure}

In this section, we evaluate the proposed approach on several benchmark tasks including image classification, semantic segmentation, and depth estimation. For image classification, we consider the popularly used dataset CIFAR100 and ImageNet1K. % and ImageNet~\cite{imagenet_cvpr09}. 
For semantic segmentation and depth estimation, we consider CityScapes~\cite{Cordts2016Cityscapes} and NYUv2~\cite{Silberman:ECCV12} respectively. 
To make fair comparison, we use exactly the same training setting and hyper-parameters for all the methods, including data pre-processing, learning rate scheduler, number of training epochs, batch size, etc.  We first demonstrate the effectiveness of \texttt{DistPro} for classification on CIFAR100. Then, we provide more analysis with a larger-scale dataset, ImageNet1K. At last, we show the results of dense prection tasks and ablation study. All experiments are performed with Tesla-V100 GPUs. 
% \begin{table}[tbhp]
%     \centering
%     \begin{tabular}{l|c|c|c}
%         \hline
%         Dataset & $L_t$ & $L_s$ & Transforms\\
%         \hline
%         CIFAR-100 & $3$ & $3$ & $N=0,1,2$\\
%         % ImageNet & $5$ & $5$ & $N=2$\\
%         ImageNet1K & $5$ & $5$ & $N=1$\\
%         CityScapes & $5$ & $5$ & $N=1$\\
%         NYUv2 & $5$ & $5$ & $N=1$\\
%         \hline
%     \end{tabular}
%     \caption{Size of the search space on each dataset. For example, on CIFAR-100, $3$ feature maps of the teacher and $3$ of the student are selected. For each pair, three transform blocks as in Figure~\ref{fig:trans} are inserted, where $N = 0,1,2$ respectively.}
%     \label{tab:searchspace}
% \end{table}
\begin{table*}[!t]
    % \centering
    \resizebox{1.0\linewidth}{!}{%
    \begin{tabular}{l|cccccc}
        % \hline
        Teacher &  WRN-40-2 & WRN-40-2 & ResNet32x4 & ResNet32x4 & ResNet56 & ResNet110\\
        Student &  WRN-16-2 & ShuffleNet-v1 & ShuffleNet-v1 & ShuffleNet-v2 & ResNet20 & ResNet32\\
        \hline
        \hline
        Teacher Acc. & 76.51 & 76.51 & 79.45 & 79.45 & 73.28 & 74.13\\
        Student Acc. & 73.26 (0.050) & 70.50 (0.360) & 70.50 (0.360) & 71.82 (0.062) & 69.06 (0.052) & 71.14 (0.061)\\
        \hline
                L2T-ww\textsuperscript{+}~\cite{jang2019learning}& - & -& 76.35 & 77.39 & - & - \\
        ReviewKD\textsuperscript{+}~\cite{chen2021distilling} & 76.20 (0.030) & 77.14 (0.015) & 76.41 (0.063) & 77.37 (0.069) & 71.89 (0.056) & 73.16 (0.029)\\
        % \hline
        Equally weighted & 75.50 (0.010) & 74.28 (0.085) & 73.54 (0.120) & 74.39 (0.119) & 70.89 (0.065) & 73.19 (0.052)\\
        Use $\alpha_T$ & 76.25 (0.034) & 77.19 (0.074) & 77.15 (0.043) & 76.64 (1.335) & 71.24 (0.014) & 73.58 (0.012)\\
        DistPro & \textbf{76.36 (0.005)} & \textbf{77.24 (0.063)} & \textbf{77.18 (0.047)} & \textbf{77.54 (0.059)} & \textbf{72.03 (0.022)} & \textbf{73.74 (0.011)}\\
        % \hline
    \end{tabular}%
    }
    \caption{Results on CIFAR100. Results are averaged over $5$ runs. Variances are in the parentheses. ``+'' represents our reproduced results. In ``Equally weighted", we do not use the searched $\alpha$ in the retrain phase. Instead, each element of $\alpha$ is uniformly set to $1 / L$, where $L$ is the length of $\alpha$. In ``Use $\alpha_T$", the finally converged $\alpha$ is used at each iteration of the retrain phase. }
    \vspace{-15pt}
    \label{tab:cifar}
\end{table*}

\subsection{Classification on CIFAR100}
\vspace{-5pt}
% In this section, We demonstrate the improvement on model accuracy provided by \texttt{DistPro} for CIFAR-100 compared to our baseline distillation method Knowledge Review~\cite{chen2021distilling}.

\noindent \textbf{Implementation details} We follow the same data pre-processing approach as in~\cite{chen2021distilling}. Similarly, we select a group of representative network architectures including ResNet~\cite{he2016deep}, WideResNet~\cite{zagoruyko2016wide}, MobileNet~\cite{howard2017mobilenets,sandler2018mobilenetv2}, and ShuffleNet~\cite{zhang2018shufflenet,ma2018shufflenet}. We follow the same training setting in~\cite{chen2021distilling} for distillation. We train the models for $240$ epochs and decay the learning rate by $0.1$ for every $30$ epochs after the first $150$ epochs. Batch size is $128$ for all the models. We train the models with the same setting five times and report the mean and variance of the accuracy on the testing set, to demonstrate the improvements are significant. As mentioned, before distillation training, we need to obtain the distillation process by searching described as below.

To search the process $\mathcal{A}$, we randomly split the original training set into two subdatasets, $80\%$ of the images for training and $20\%$ for validating. We run the search for $40$ epochs and decay the learning rate for $\theta$ (parameters of the student) by $0.1$ at epoch $10$, $20$, and $30$. The learning rate for $\alpha$ is set to $0.05$.
% On ImageNet, we run the search for $5$ epochs and decay the learning rate by $0.1$ at epoch $2$, $3$, and $4$.
Following the settings in~\cite{chen2021distilling}, we did not use all feature maps for knowledge distillation but only the ones after each downsampling stage. Transform blocks are used to transform the student feature map. Fig.~\ref{fig:trans} shows the block architecture. The size of the pathways is 27, as we use 3 transform blocks and 9 connections. To make fair comparison, we follow the same HCL distillation loss in~\cite{chen2021distilling}. Once the the process is obtained, in order to align the searched distillation process, linear interpolation is used to expand the process of $\mathcal{A}$ from 40 epoch (search stage) to 240 epochs (retrain stage) during KD.

\noindent\textbf{Results} The quantitative results are summarized in Table~\ref{tab:cifar}. First two rows show the architecture of teacher and student network and their accuracy without distillation correspondingly. In the line of ``ReviewKD\textsuperscript{+}'', we list the results reproduced using the code released by the author~\footnote{\url{https://github.com/dvlab-research/ReviewKD}}, notice it could be a bit different with reported due to randomness in data loading. To demonstrate learning the $\alpha$ is essential, we first assign equally weighted $\alpha$ to all the pathways. As shown in line ``Equally weighted'', the results are worse than ReviewKD, indicating the selected pathways from ReviewKD are useful. For ablation, we first adopt the learned $\alpha_{T}$ at the end of the process. As shown in ``Use $\alpha_T$'', it outperforms ReviewKD in multiple settings. The last row shows the results adopting the learned process $\mathcal{A}$, which outperforms ReviewKD significantly based on our derived variations. %We believe this is because at the search stage, the student network weights $\omega$ is evolving with a similar behavior as the distillation  stage. Therefore, this KD scheme with the process is further improved.

\begin{figure}[t]
    \scalebox{0.9}{
    \begin{minipage}{.45\linewidth}
    \centering
    \includegraphics[width=\textwidth]{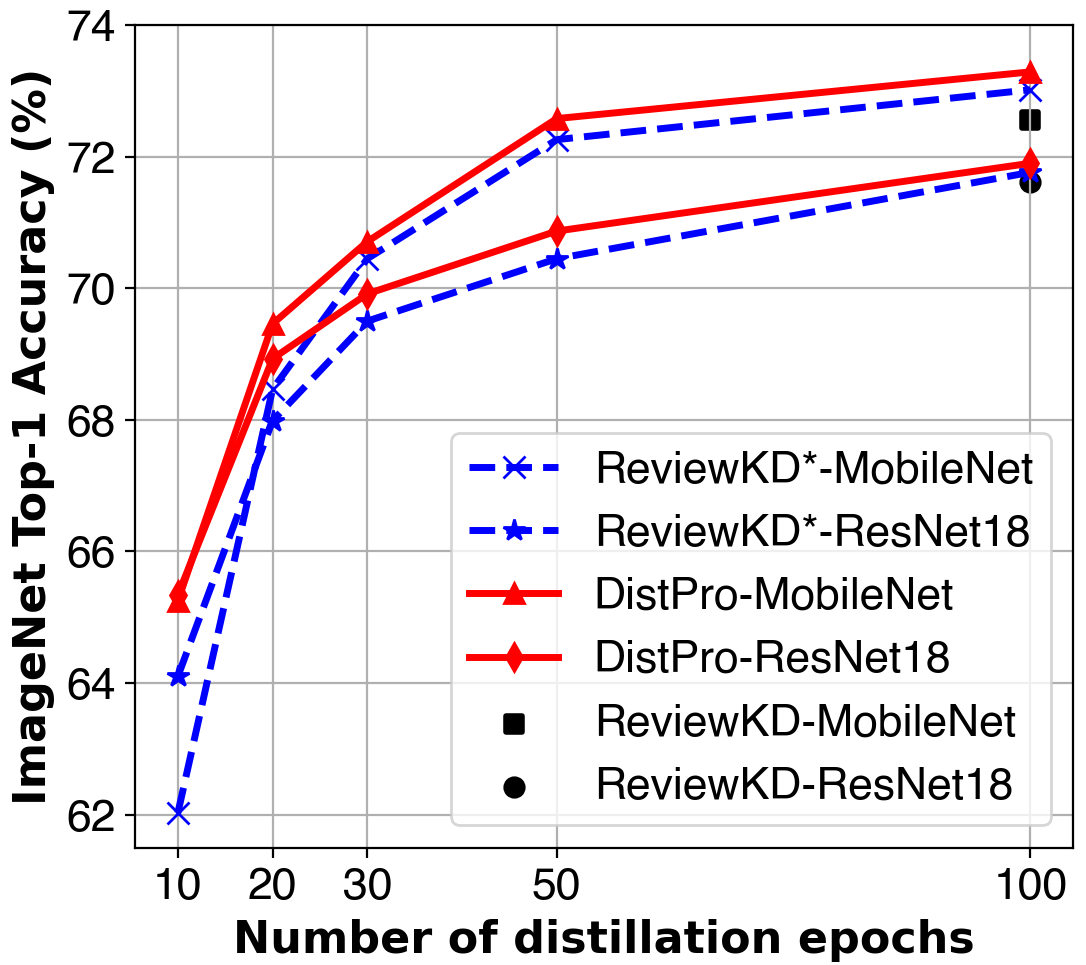}
\vspace{-20pt}
    \caption{Comparison study on fast distillation on ImageNet1K. }
    \label{fig:speed}
    \end{minipage}
    }
    \begin{minipage}{.55\linewidth}
        \resizebox{1.0\linewidth}{!}{%
    \begin{tabular}{l|c|cc}
         & Network            & ReviewKD\cite{chen2021distilling} & DistPro(Ours)\\
         \hline
         \hline
Top-1    & \multirow{2}{*}{MobileNet} & 72.56    & 72.54   \\
\#Epochs &                      & 100      & \textbf{50}      \\
\hline
Top-1    & \multirow{2}{*}{ResNet18} & 71.61    & 71.59   \\
\#Epochs &                      & 100      & \textbf{65}      \\
\hline
Top-1    & \multirow{2}{*}{DeiT} & 73.44    & 73.41   \\
\#Epochs &                      & 150      & \textbf{100}    
\end{tabular}%
}
\captionof{table}{Comparison study on numbers of distillation epochs to achieve same top-1 accuracy on ImageNet1K (lower the better). }
\label{tab:speed}
    \end{minipage}
    \vspace{-10pt}
\end{figure}

\vspace{-8pt}
\subsection{Classification on ImageNet1K} 
\label{sec:exp_imagenet}
\vspace{-5pt}
% In this section, we evaluate our proposed method \texttt{DistPro} on ImageNet1K. \texttt{DistPro} can not only outperform the existing KD methods even with marginal improvement, but also reduce distillation time significantly.

\noindent \textbf{Implementation Details} We follow the search strategy used in CIFAR100 and training configurations in \cite{chen2021distilling} with a batch size of 512 (4GPUs are used). 
The selected student/teacher networks are MobileNet/ResNet50, ResNet18/ResNet34 and DeiT-tiny~\cite{touvron2021deit}/ViT-B~\cite{dosovitskiy2020vit}. We adopt a different architecture of transform block for DeiT.  % We first search the distillation process $\mathcal{A}$ for ImageNet1K and later retrain KD with the searched $\mathcal{A}$. 
At searching stage, we search $\mathcal{A}$ with Tiny-ImageNet~\cite{le2015tiny} for 20 epochs. %which can result in lower search cost compared to that with full-size ImageNet. 
We adopt cosine lr scheduler for learning the student network. % and derive our experimental results in the table with cosine scheduler. 
% More details can be found in ablation study in Sec.\ref{sec:ablation}. 
For search space, 1 transform block is used, while we consider 5 feature maps in the networks, and build 15 pathways by removing some pathways following~\cite{chen2021distilling} due to limited GPU memory.  % We then use the transform block to build the path way between the feature maps of teacher and student network. 
% For all three settings, we  for both student and teacher. Follow the connecting strategy used in CIFAR100, we at last build . 
At KD stage, we train the network for various epochs with initial learning rate of 0.1, and cosine scheduler. It takes 4 GPU hours for searching and ~80 GPU hours for distill with 100 epochs.
More details (ViT transform block, selected pathways etc) can be found in supplementary due to limited space.

\noindent\textbf{Fast Distillation.} From prior works~\cite{shen2021fast}, KD is used to accelerate the network learning especially with large dataset. This is due to the fact that its accuracy can be increased using only Ground Truth labels if the network is trained with a large number of epochs\cite{hoffer2017train}. Previous works~\cite{chen2021distilling} only check the results stopping at 100 epochs. Here, we argue that it is also important to evaluate the results with less training cost, which is another important index for evaluating KD methods. This will be practically useful for the applications requiring fast learning of the network. 

Here, we show \texttt{DistPro} reaches much better trade-off between training cost and accuracy. 
In Fig.~\ref{fig:speed}, %shows the comparison on performance with various distillation epochs. As shown in the figure, 
\texttt{DistPro} with different network architectures (red curves) outperforms ReviewKD with the same setting (dashed blue curves) at all proposed training epochs. % It is noted that the comparison is made based on the same training configuration with cosine learning rate scheduler. 
The performance gain is larger when less training cost is required, \textit{e.g.} it is 3.22\% on MobileNet with 10 epochs (65.25\% vs 62.03\%), while decreased to 0.14\% with 100 epochs, which is still a decent improvement. % In details, the improvement margin for MobileNet is reduced to 0.99\% with 20 epochs, 0.33\% with 30 epochs, 0.28\% with 50 epochs, and 0.14\% with 100 epochs. 
Similar results are observed with ResNet18. Note here for all results with various epochs, we adopt the same learning process $\mathcal{A}$, where the time cost can be ignored comparing to that of KD. % The searched process for three settings can be found in supplementary.
In Tab.~\ref{tab:speed}, we show the number of epochs saved by \texttt{DistPro} when achieving the same accuracy using ReviewKD~\cite{chen2021distilling}. For instance, in training MobileNet, \texttt{DistPro} only use 50 epochs to achieve 72.54\%, which is comparable to ReviewKD trained with 100 epochs (72.56\%), yielding 2x acceleration. Similar acceleration is also observed with ResNet18 and DeiT. %, demonstrating the effectiveness of  % For ResNet18, \texttt{DistPro} reaches 71.59\% with 65 epochs (ReviewKD~\cite{chen2021distilling}:71.61\% with 100 epochs), yielding 35\% 

\begin{table}[t]
\begin{tabular}{c|cc|cc|c}
Setting & Search dataset & Retrain dataset & Teacher         & Student        & Top-1 (\%) \\
\hline
\multirow{2}{*}{(a)}    & Tiny-ImageNet     & ImageNet1K       & ResNet50        & MobileNet      & 73.26      \\
        & CIFAR100      & ImageNet1K     & ResNet50        & MobileNet      & 73.20      \\
        
        % \hline
        \midrule
Setting & Search teacher & Search student  & Retrain teacher & Retran student & Top-1 (\%) \\
        \hline
\multirow{2}{*}{(b)}     & ResNet34       & ResNet18        & ResNet34        & ResNet18       & 71.89      \\
        & ResNet50       & MobileNet       & ResNet34        & ResNet18       & 71.87     
\end{tabular}
\caption{Results of transferring searched $\mathcal{A}$ cross search-retrain datasets and search-retrain networks performed on ImageNet1K with 100 epochs. Top-1 accuracy on validation set is reported. }
\vspace{-20pt}
\label{tab:transfer}
\end{table}

\noindent \textbf{Transfer $\mathcal{A}$ } Here, we study whether learned $\mathcal{A}$ can be transferred across datasets and similar architectures. Tab.~\ref{tab:transfer} shows the results. In setting (a), we resize the image in CIFAR100 to $224\times 224$, and do the process search on CIFAR100, where the search cost is only 2 GPU hours. As shown at last column, it only downgrades accuracy by 0.06\%, which is comparable with full results searched with ImageNet, demonstrating the process could be transferred across datasets. In setting (b), we adopt the searched process with student/teacher networks of MobileNet/ResNet50 to networks of ResNet18/ResNet34 since they have same feature pathways at similar corresponding layers. As shown, the results are closed, demonstrating the process could be transferred when similar pathways exist.
% We conduct two groups of experiments: 1) search \textbf{$\mathcal{A}$} on CIFAR100 then retrain on ImageNet1K 2) search \textbf{$\mathcal{A}$} on Tiny-ImageNet where ResNet50 is served as teacher and MobileNet as student then retrain on ImageNet1K using ResNet18 as student and ResNet34 as teacher. 
% As shown in the table, with the same distillation networks, \texttt{DistPro} is able to achieve closed performance cross different search and retrain datasets yielding 73.26\% searched on Tiny-ImageNet and 73.20\% searched on CIFAR100 respectively. We also demonstrate that the learned process $\mathcal{A}$ is transferable across different networks. The process $\mathcal{A}$ searched on ResNet50 and MobileNet can achieve top1 of 71.87\% when retrained on ImageNet1K yielding closed performance of that searched and retrained both on ResNet34 and ResNet18 (73.26\%).  However, it should be noted that the amount of path ways should remain the same. For example, there are 15 path ways between ResNet50 and MobileNet and the therefore the objective pair of teacher (ResNet34) and student (ResNet18) should also keep 15 path ways.

\begin{table}[b]
    \resizebox{1.0\linewidth}{!}{%
\begin{tabular}{c|c|cc|cccccc}
Setting              &   Acc (\%)    & Teacher & Student & OFD\cite{heo2019comprehensive}  & LONDON\cite{shang2021london} & SCKD\cite{zhu2021sckd} & ReviewKD\cite{chen2021distilling} & ReviewKD* & DistPro\\
\hline
\hline
\multirow{2}{*}{(a)} & Top1 & 76.16   & 68.87   & 71.25 & 72.36  & 72.4 & 72.56    & 73.12     & \textbf{73.26}          \\
                     & Top5 & 93.86   & 88.76   & 90.34 & 91.03  & -    & 91.00    & 91.22     & \textbf{91.27}          \\
                     \hline
\multirow{2}{*}{(b)} & Top1 & 73.31   & 69.75   & 70.81 & -      & 71.3 & 71.61    & 71.76     & \textbf{71.89}          \\
                     & Top5 & 91.42   & 89.07   & 89.98 & -      & -    & 90.51    & 90.67     & \textbf{90.76}          \\
                     \hline
(c)                  & Top-1 & 85.1    & 73.41    & -     & -      & -    & -        & 73.44     & \textbf{73.51}         
\end{tabular}%
}

\caption{Comparison study on ImageNet1K. Settings are (a) teacher: ResNet-50, student: MobileNet; (b) teacher: ResNet-34, student: ResNet-18; (c) teacher: ViT-B, student: DeiT-tiny. ReviewKD* denotes our reproduced experimental results with cosine learning rate scheduler. }
\vspace{-20pt}
\label{tab:imagenet}
\end{table}

\noindent\textbf{Best Results.} Finally, Tab.\ref{tab:imagenet} shows the quantitative results comparing with various SoTA baselines. As mentioned, we adopt cosine scheduler while ReviewKD~\cite{chen2021distilling} adopts step scheduler. To make it fair, we retrain ReviewKD with cosine scheduler, and list the results in the column of ReviewKD*. As shown in the table, for three distillation settings, \texttt{DistPro} outperform the existing methods achieving top-1 accuracy of 73.26\% for MobileNet, 71.89\% for ResNet18 and 73.51\% for DeiT respectively, yielding new SoTA results for all these networks. % In details, \texttt{DistPro} can provide a large margin of 0.7\% compared to ReviewKD~\cite{chen2021distilling} for MobileNet, while the improvement is less significant for other networks. This may result from the cosine scheduler, as it can also raise the score of ReviewKD from 72.56\% to 73.12\% (ReviewKD*). While it is less beneficial to ResNet18, as the improvement of ResNet18 is only 0.15\%.  There is no improvement by changing to cosine scheduler for Deit as cosine is the original scheduler~\cite{touvron2021deit}.

% Another finding is that LATTE produce larger improve margin when teacher and student network has more difference in architecture, \eg from ``ResNet32x4'' to ``ShuffleNet-v1'', LATTE produces $+0.77$ improvements while 
% Second, as shown in Figure~\ref{fig:imagenet}, with the optimized varying teaching scheme, the training is faster than using the fixed, hand-crafted teaching scheme in knowledge review. However, the final performance of our approach on ImageNet does not beat that of knowledge review. The reason might be that we did not run enough iterations in the search phase on ImageNet.

% \begin{figure}[tbhp]
%     \centering
%     \includegraphics[width=0.49\linewidth]{figures/ImageNet-1.pdf}
%     \includegraphics[width=0.49\linewidth]{figures/ImageNet-2.pdf}
%     \caption{Accuracy on ImageNet. As shown in the figure, with our method, the training is faster.}
%     \label{fig:imagenet}
% \end{figure}

% \begin{table}[]

% \end{table}

\vspace{-5pt}
\subsection{Dense prediction tasks}
\label{subsec:dense}
\vspace{-5pt}
\noindent\textbf{CityScapes} is a popular semantic segmentation dataset with pixel class labels~\cite{Cordts2016Cityscapes}. We compare to a SoTA segmentation KD method called IFVD~\cite{wang2020intra} and adopt their released code~\footnote{https://github.com/YukangWang/IFVD}. IFVD is a response-based KD method, therefore can be combined with feature-based distillation method including ReviewKD and \texttt{DistPro}, which are shown as ``+ReviewKD'' and ``+\texttt{DistPro}'' respectively in Table~\ref{tab:cityscapes}.  We adopt student of MobileNet-v2 to compare against another SoTA results from SCKD~\cite{zhu2021sckd}. Please note here for simplicity and numerical stability, we disable adversarial loss of the original IFVD. From the table, ``+\texttt{DistPro}'' outperforms all competing methods.  % We report the mIoU on the validation set in Table~\ref{tab:cityscapes}. Please note that the original IFVD code did not run due to numerical issues, and we resorted to disabling the adversarial training loss in IFVD.

\noindent\textbf{NYUv2} is a dataset widely used for depth estimation~\cite{Silberman:ECCV12}. The experiments are based on the code~\footnote{https://github.com/fangchangma/sparse-to-dense} released in S2D~\cite{ma2018sparse}. We compare \texttt{DistPro} with the plain KD, where we treat teacher output as ground truth, and ReviewKD with intermediate feature maps. Root mean squared errors (RMSE) are summarized in Table~\ref{tab:nyu}, and it shows that \texttt{DistPro} is also beneficial. In the future, we will explore more to verify its generalization ability.

\begin{table}[t]
    \scalebox{0.95}{
    \begin{minipage}{.33\linewidth}
    %   \centering
      \resizebox{1.0\linewidth}{!}{%
        \begin{tabular}{l|l}
        % \hline
        Teacher & ResNet101\\
        Student & MobileNet-v2\\
        \hline
        \hline
        Baseline & 66.92 (0.00721)\\
        IFVD~\cite{wang2020intra} & 68.31 (0.00264)\\
        SCKD~\cite{zhu2021sckd} & 68.25 (0.00307) \\
        +ReviewKD~\cite{chen2021distilling} & 69.03 (0.00373)\\
        +Equally Weighted $\alpha$ & 68.49 (0.00793)\\
        \hline
        +\texttt{DistPro} & \textbf{69.12 (0.00462)}\\
        % \hline
        \end{tabular}
        }
      \caption{mIoU (\%) on CityScapes (higher the better). Results are averaged over $5$ runs. Standard deviation in the parentheses. }
      \label{tab:cityscapes}
    \end{minipage}}%
    \quad\quad
    \scalebox{0.95}{
    \begin{minipage}{0.33\linewidth}
    %   \centering
        \resizebox{1.0\linewidth}{!}{%
        \begin{tabular}{l|l}
        % \hline
        Teacher & ResNet50\\
        Student & ResNet18\\
        \hline
        \hline
        Baseline & 0.2032\\
        KD & 0.2045\\
        ReviewKD~\cite{chen2021distilling} & 0.1983\\
        Equally Weighted $\alpha$ & 0.2030\\
        \hline
        \texttt{DistPro} & \textbf{0.1972}\\
        % \hline
        \end{tabular}
        }
    \caption{Estimation error on NYUv2 (lower the better). \\\hfill\\\hfill\hfill\hfill\hfill}
    \label{tab:nyu}
    \end{minipage}}
        \scalebox{0.95}{
    \begin{minipage}{0.3\linewidth}
      \resizebox{1.0\linewidth}{!}{%
       \begin{tabular}{c|c}
        Normalization & Acc. \\
        \hline
        \hline
        $\texttt{softmax}(\texttt{concat}(\tilde{\alpha}, 1))$ & \textbf{79.78}\\
        % \hline
        $\texttt{softmax}(\tilde{\alpha})$ & 79.41\\
        % \hline
        $\frac{\tilde{\alpha}}{\|\tilde{\alpha}\|_1 + 1}$ & 79.47\\
        % \hline
        $\frac{\tilde{\alpha}}{\|\tilde{\alpha}\|_1}$ & 79.20\\
        % \hline
        $\texttt{sigmoid}(\tilde{\alpha})$ & 78.80\\
        % \hline
    \end{tabular}
    }
   \captionof{table}{Performance of different normalization methods on CIFRA-100. $\texttt{concat}(\tilde{\alpha}, 1)$ concatenate $\tilde{\alpha}$ with a scalar $1$. Results are averaged over $5$ runs.}
    \label{tab:normalization}
      
    \end{minipage}}
    \vspace{-25pt}
% \caption{Global caption}
\end{table}

% \paragraph{} The above results verify that the proposed approach is also beneficial for dense prediction tasks.

\vspace{-5pt}
\subsection{Ablation study}
\label{sec:ablation}
\vspace{-5pt}

\noindent\textbf{Is it always beneficial to transfer knowledge only from lower-level feature maps to higher-level feature maps?} In Table~6 of ~\cite{chen2021distilling}, the authors conducted a group of experiments on CIFAR100, which show that only pathways from lower-level feature maps in teacher to higher-level feature maps in student are beneficial. However, we conducted similar experiments on CityScapes, the results did not support this claim. Specifically, using pathways from the last feature map to the first three feature maps results in mIoU $73.9\%$ on ResNet18, while using pathways from all lower-level feature maps to higher level ones as in~\cite{chen2021distilling} results in $73.5\%$. This suggests that the optimal teaching scheme need to be searched w.r.t larger space. Also, the results in Table~\ref{tab:cityscapes} also show that our searched process is better than the hand-crafted one. % More interestingly, as will be shown later, the searched scheme suggests that the student should learn from different teacher feature maps at different stages of the training process.

\begin{figure}[t]
    \centering
    \includegraphics[width=0.32\textwidth]{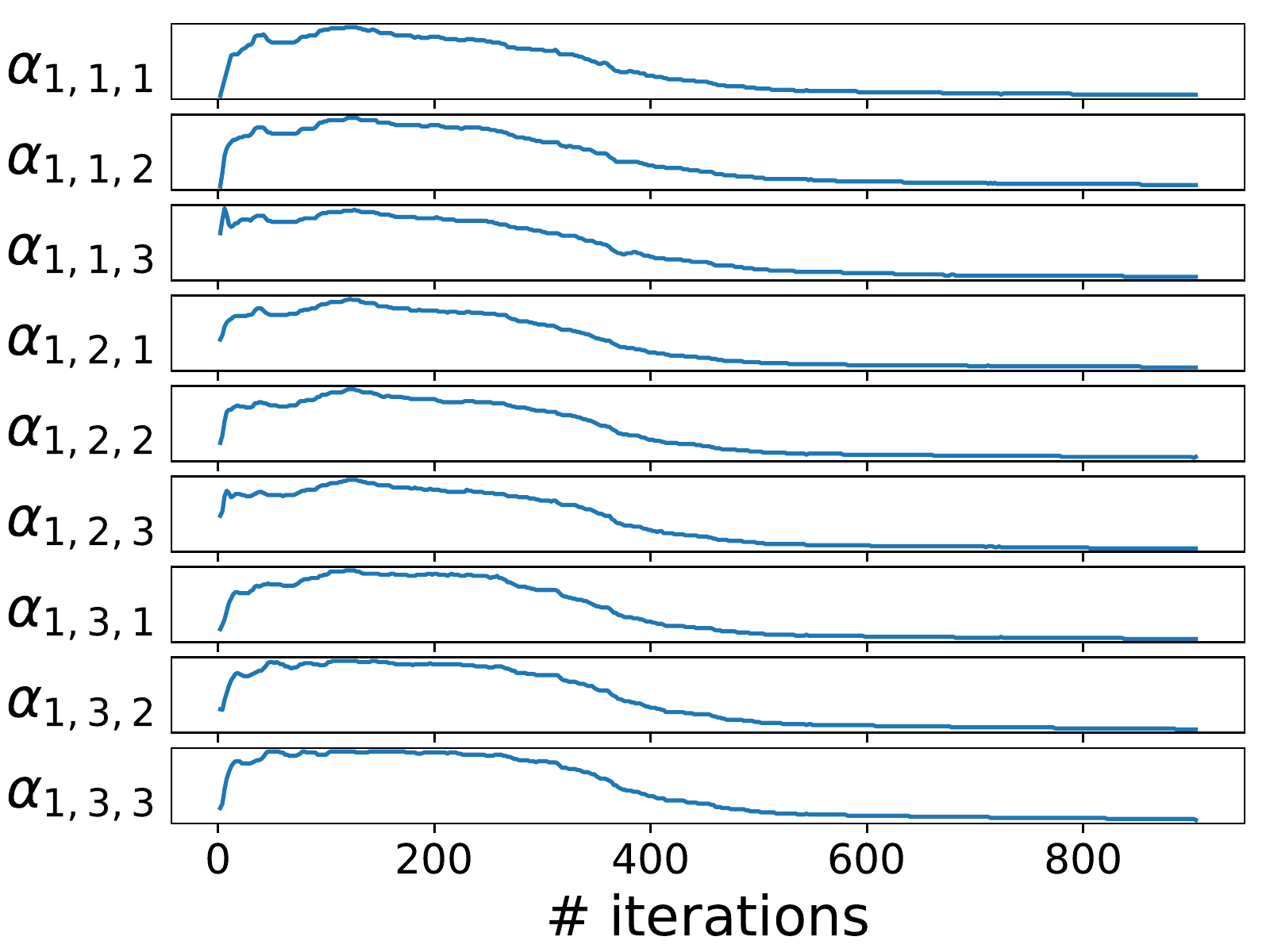}
    \includegraphics[width=0.32\textwidth]{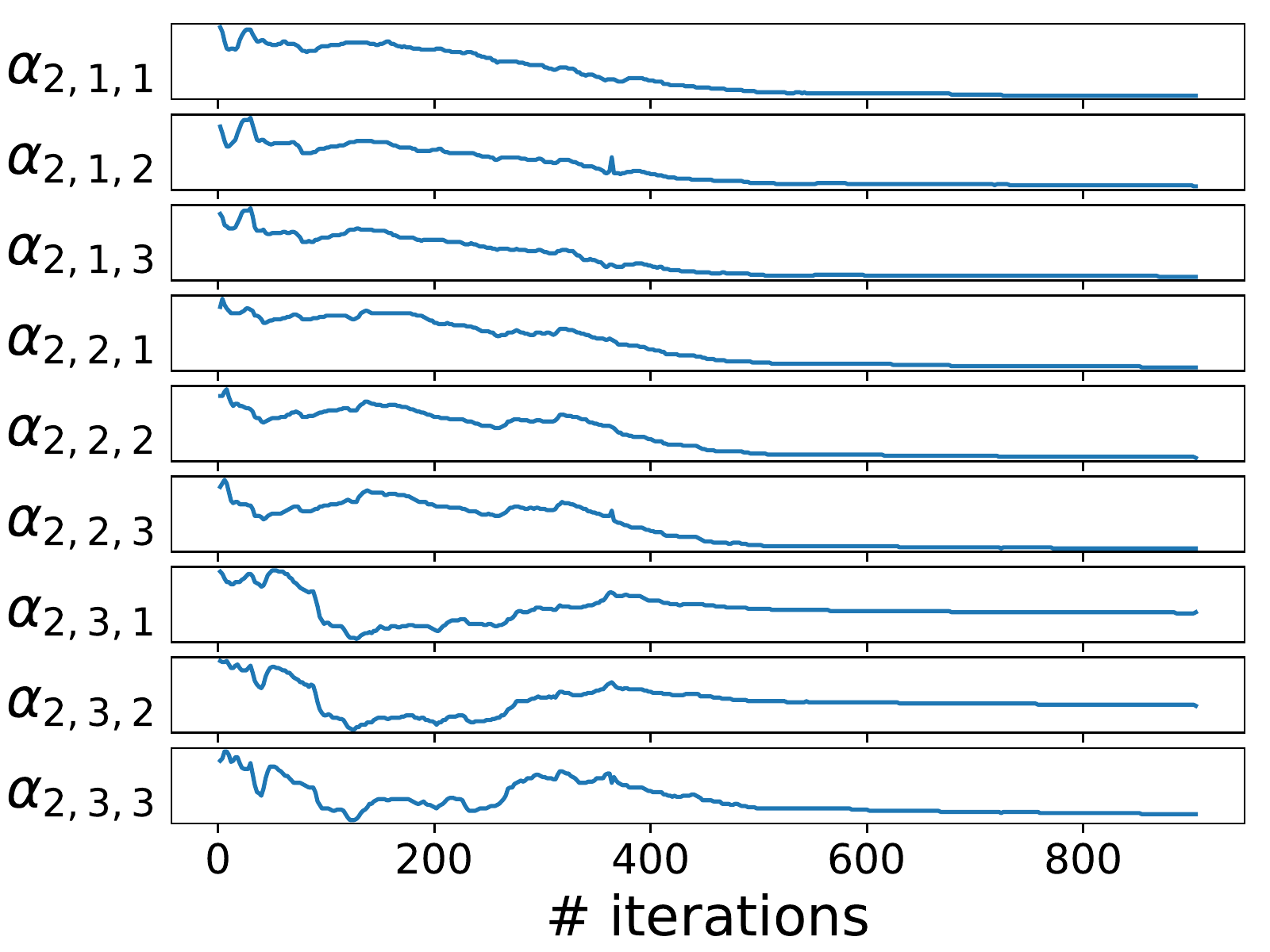}
    \includegraphics[width=0.32\textwidth]{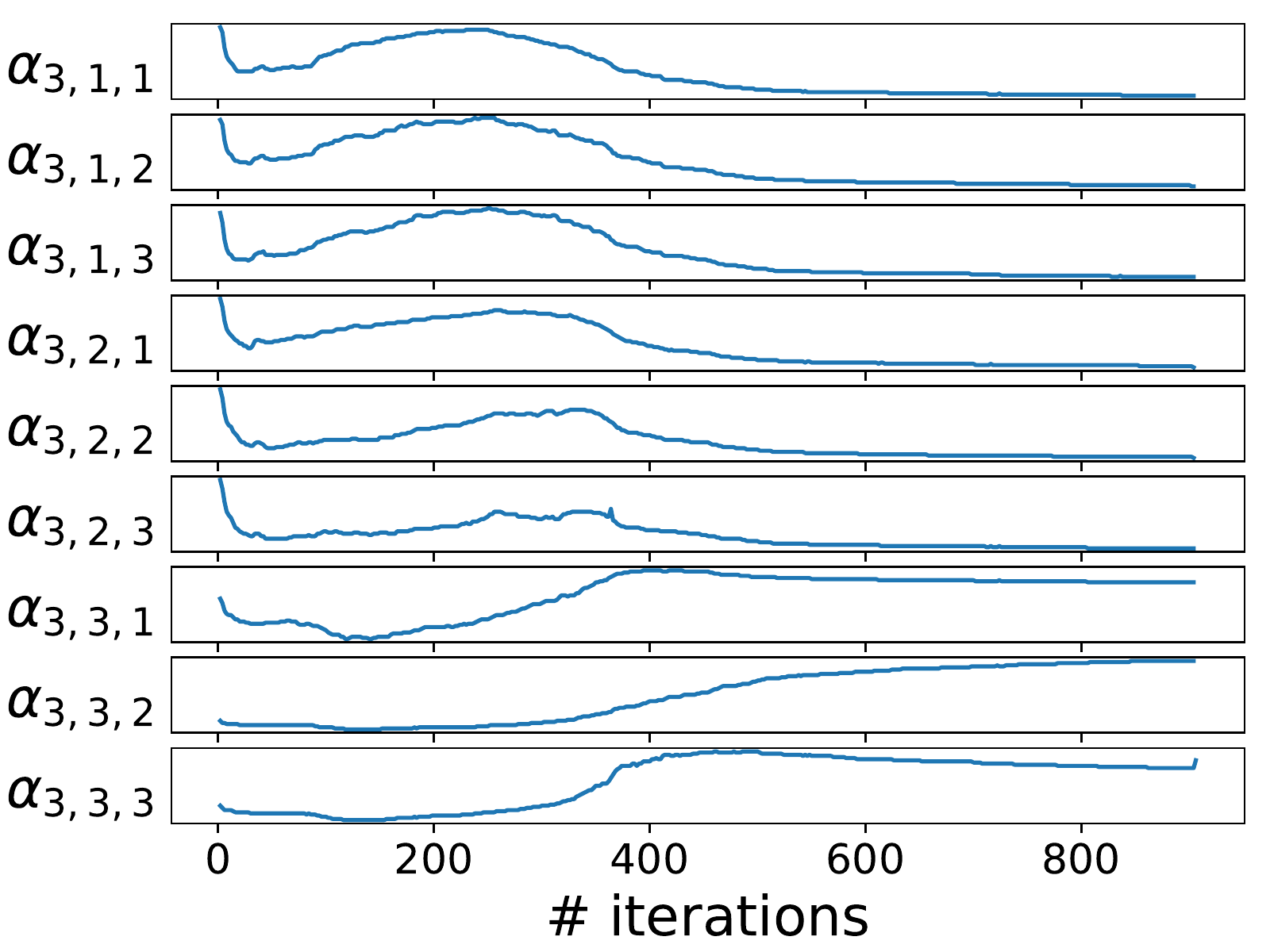}
    \vspace{-10pt}
    \caption{Searched distillation process performed on CIFAR100 with WRN-40-2 as teacher and WRN-16-2 as student. $x$-axis is the number of iterations. The left image contains $\alpha[1,:,:]$, i.e., all the elements corresponding to the lowest-level feature map of the teacher. The middle one corresponds to $\alpha[2,:,:]$, and the right one corresponds to $\alpha[3,:,:]$.}
    \vspace{-15pt}
    \label{fig:alphas}
\end{figure}

\noindent \textbf{Intuition of process $\mathcal{A}$} %As mentioned before, one design choice in the proposed approach is how to make use of the $\alpha$ generated in the search phase. 
In Figure~\ref{fig:alphas}, we show how the learned sample of $\mathcal{A}$ changes with time in search stage of distilling WRN-40-2 to WRN-16-2. Similar observations are found in other settings. The figure indicates that at the early stage of the training, the optimized teaching scheme focuses on transferring knowledge from low-level feature maps of the teacher to the student. As training goes, the optimized teaching scheme gradually moves on to the higher-level feature maps of the teacher. Intuitively, high-level feature maps encode highly abstracted information of the input image and thus are harder to learn from compared to the low-level feature maps. \texttt{DistPro} is able to automatically find a teaching scheme to make use of this intuition. %This suggests that $\alpha$ as a function of time encodes much more information than the finally converged $\alpha$, and thus we claim that we should make use of all the $\alpha$'s generated in the search phase. The results of retraining using only the finally converged $\alpha$ is summarized in Table~\ref{tab:cifar} as ``Use final $\alpha$, which clearly verify our claim.

\noindent \textbf{Normalization of $\alpha_t$} As mentioned before, we apply normalization to $\alpha_t$ (\eqref{eq:alphatrain}) for numerical stability after $2_{nd}$ gradient approximation. In this study, we evaluate several  normalization strategies. Let the unnormalized parameters be $\tilde{\alpha_t}$ and the normalized ones be $\alpha_t$. In the following, we view $\alpha_t$ as a vector of length $L$. The experiments are conducted on CIFAR100 with WRN-28-4 as teacher and WRN-16-4 as student. The results are shown in Table~\ref{tab:normalization}, and the proposed biased softmax normalization outperforms others.  Intuitively, due to the appended scalar $1$, the value of $\tilde{\alpha_t}$ can be compared against $1$, and a constant value yields a label smoothing~\cite{muller2019does} effect for the distribution.

% \noindent \textbf{Distillation curve} Fig.~\ref{fig:curve} shows the comparison of learning curves of ImageNet1K for ReviewKD and \texttt{DistPro}. As shown in the figure, \texttt{DistPro} can distill much faster at earlier epoch compared to ReviewKD for both trained with 20 and 50 epochs.

% CVPR targets.  
% 1) CIFAR 100 all experimental settings tunning. 
% 2） imagenet  parameter tunning. 
% 3)  dense prediction: different network settings: resnet, shufflenet etc. res18 no improvement. 
% 4） larger sample space. 
\vspace{-5pt}
\section{Conclusion}
\vspace{-8pt}
In this paper, we take a step towards the problem of finding the optimal KD scheme given a pair of wanted student network and learned teacher network under a vision task. 
Specifically, we setup a searching space by building pathways between the two networks and assigning a stochastic distillation process along each path pathway.
We propose a meta-learning framework, \texttt{DistPro}, to learn these processes efficiently, and find effective ones to perform KD with intermediate features. We demonstrate its benefits over image classification, and dense predictions such as image segmentation and depth estimation.
We hope our method could inspire the field of KD to further expand the scope, and its cooperation with other techniques such as NAS and hyper-parameter tuning.

\clearpage
% ---- Bibliography ----
%
% BibTeX users should specify bibliography style 'splncs04'.
% References will then be sorted and formatted in the correct style.
%
\bibliographystyle{splncs04}
\bibliography{egbib}
\end{document}